\documentclass[]{journal}  

\usepackage{amsmath,amsfonts,amssymb}
\usepackage{algorithm2e}
\usepackage{graphicx}
\usepackage{todonotes}

\usepackage[colorlinks=true, allcolors=blue]{hyperref}
\usepackage{float}
\usepackage{comment}
\usepackage{booktabs}
\usepackage{caption}
\usepackage{multirow}
\usepackage{blindtext}
\usepackage{multicol}
\usepackage{amsmath, amssymb}
\usepackage{tabularx}
\usepackage[margin=0.9in]{geometry}
\usepackage{}
\title{Beyond Images: An Integrative Multi-modal Approach to Chest X-Ray Report Generation}

\author[a,b]{Nurbanu Aksoy}
\author[c]{Serge Sharoff}
\author[f]{Selcuk Baser}
\author[a,b,*]{Nishant Ravikumar}
\author[a,d,e,*]{Alejandro F Frangi}
\affil[a]{CISTIB Centre for Computational Imaging and Simulation Technologies in Biomedicine}
\affil[b]{School of Computing, University of Leeds, Leeds, UK}
\affil[c]{School of Languages, Cultures and Societies, University of Leeds, Leeds, UK}
\affil[d]{School of Health Sciences, Faculty of Biology, Medicine and Health, The University of Manchester}
\affil[e]{School of Computer Science, Faculty of Science and Engineering, The University of Manchester}
\affil[f] {Department of Radiology, Kastamonu Training and Research Hospital, Kastamonu, Turkey}

\affil[*]{Indicates joint last authors}

\excludecomment{mycomment}
 
\begin{document} 
\maketitle

\begin{abstract}
Image-to-text radiology report generation aims to automatically produce radiology reports that describe the findings in medical images. Most existing methods focus solely on the image data, disregarding the other patient information accessible to radiologists. In this paper, we present a novel multi-modal deep neural network framework for generating chest X-rays reports by integrating  structured patient data, such as vital signs and symptoms, alongside unstructured clinical notes.We introduce a conditioned cross-multi-head attention module to fuse these heterogeneous data modalities, bridging the semantic gap between visual and textual data. Experiments demonstrate substantial improvements from using additional modalities compared to relying on images alone. Notably, our model achieves the highest reported performance on the ROUGE-L metric compared to relevant state-of-the-art models in the literature. Furthermore, we employed both human evaluation and clinical semantic similarity measurement (Bio-ClinicalBERT Score\cite{alsentzer2019publicly}) alongside word-overlap metrics to improve the depth of quantitative analysis. A human evaluation, conducted by a board-certified radiologist, confirms the model's accuracy in identifying high-level findings,  however, it also highlights that more improvement is needed to capture nuanced details and clinical context.
\end{abstract}


\begin{multicols}{2}

\section{Introduction}
\label{sec:intro}  
The use of medical imaging is widespread across various branches of health sciences for the purpose of diagnosing diseases, developing effective treatment plans, providing patient care, and predicting disease outcomes. Radiologists are responsible for interpreting the medical images and creating a full-text radiology report that is based on their findings along with other relevant clinical data and information, such as patient demographics, symptoms, and pre-existing/existing medical conditions. These reports must be complete, accurate, and produced in a short amount of time while adhering to a specific format. In clinical settings, chest X-rays (CXR) are the most commonly used medical imaging techniques and are usually the first step in evaluating patients for various lung diseases. The reports generated from CXR examinations typically include the radiologists' observations categorised as "findings" and "impressions" and indicate normal and abnormal features in the images. Composing these detailed reports requires a significant amount of knowledge and experience and can be time-consuming and prone to errors. By providing radiologists with a baseline analysis to validate and amend as needed, automation can reduce repetitive workflows. This would allow radiologists to focus their expertise on higher-level clinical thinking and quality assurance.
\\
\\
In the field of medical imaging informatics, previous studies have developed techniques to automate the generation of radiology reports \cite{yang2020automatic,singh2021show}. The majority of current deep learning approaches use networks that feature a convolutional encoder and recurrent \cite{jing2017automatic,xue2018multimodal,yuan2019automatic} or transformer decoder \cite{chen2020generating,nooralahzadeh2021progressive}, which were originally designed for the task of image captioning.  Although these two tasks share similarities in terms of input and output modalities, there are some key differences. Radiology reports are in the form of detailed paragraphs rather than brief captions, and they must be comprehensive and include specific medical details. Additionally, interpreting medical images can be challenging due to subtle variations in the image and report and also generating a description for a medical image often necessitates supplementary information beyond what is visible in the image. For instance, in certain cases, while similarities in medical imaging between males and females are nearly identical in terms of visual patterns, differences in patient demographics have a noteworthy clinical impact on the assessment and diagnosis. However, current report generation methods for CXRs solely consider the radiology image as input and disregard the non-imaging information that radiologists have access to during image interpretation. Only a limited number of studies integrate additional data into the network such as medical concepts, high-level contexts or categories of the images/reports. While these methods have shown some level of success, they mainly focus on enhancing the model with data derived from existing semantics rather than supplementing the training context with additional data. Furthermore, as CXR images are 2D projections of 3D objects, important/relevant information is lost, leading to semantic gaps in the data available for learning by networks or algorithms. Therefore, we hypothesise that combining multiple data sources that provide different perspectives on the patient's condition, is beneficial for generating more informative and accurate reports (using data-driven learning-based approaches), compared to using CXR images alone.

The rest of the paper is organised as follows: Section 2 presents relevant existing studies and strategies, and Section 3 introduces the data and outlines the proposed methodology. Section 4 provides details on the implementation and presents both quantitative and qualitative results. Finally, Section 5 conducts a discussion and draws a series of conclusions regarding the study.
\section{Relevant Literature}
\textbf{Image Captioning Methodologies}
\\
Visual Captioning, also known as Image Captioning, is a popular task that involves generating descriptive natural language captions for images. It requires the integration of computer vision and natural language processing, drawing significant attention from the artificial intelligence community. Deep learning techniques, particularly encoder-decoder models using Convolutional Neural Networks (CNNs) and Recurrent Neural Networks (RNNs), have been commonly employed for this task \cite{you2016image, anderson2018bottom}. However, these RNN-based models suffer from the issue of vanishing and exploding gradients due to limited access to previous inputs, affecting their performance \cite{ghandi2023deep}. To address this limitation, recent research has shifted towards utilising transformer-based architectures\cite{suresh2022image}, originally successful in the field of natural language processing (NLP). Transformers leverage self-attention mechanisms, allowing for better parallelisation and learning relationships between words in a sequence\cite{vaswani2017attention}. Unlike RNNs, transformers do not rely on recurrence, enabling faster and more effective learning by including more context in the network.
\begin{figure}[H]
\begin{center}
\begin{tabular}{c} 
\includegraphics[width=0.4\textwidth]{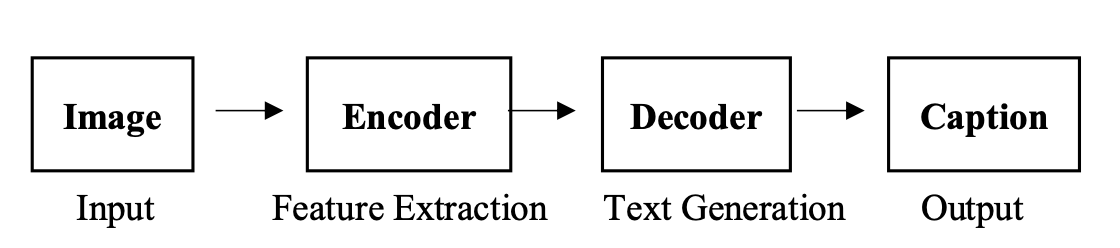}
\end{tabular}
\end{center}
\caption[results] 
{ \label{fig:results} 
Generalised image-to-text framework.}
\end{figure} 
The transformer-based encoder-decoder architecture that is most commonly used for image captioning consists of three main components: a model for extracting visual features, a transformer-based encoder, and a transformer-based decoder. To extract high-level features, pre-trained CNN models are usually employed. However, in this approach, the output of the visual model is passed through a transformer-based encoder to map the visual features and produce a sequence of image representations. The transformer-based decoder then takes in the encoder's output to generate a corresponding caption for the given image. One model, called Captioning Transformer (CT)\cite{zhu2018captioning}, utilised a ResNeXt\cite{xie2017aggregated} CNN model as an encoder and a Transformer as a decoder. Another study\cite{zhang2019image} used a Transformer as the decoder along with the ResNet CNN model and improved the network with a combination of spatial and adaptive attention. Another study\cite{li2019entangled} enhanced the vanilla Transformer architecture with Entangled Attention (ETA) and Gated Bilateral Controller (GBC), enabling the processing of semantic and visual concepts concurrently. A Meshed-Memory Transformer model\cite{cornia2020meshed} was introduced that is a fully-attentive model including a Memory-Augmented Encoder enriched with learnable keys and values and a gating mechanism for mesh connectivity. The model also includes a Meshed Decoder that connects all encoding layers. A separate study proposed the Caption TransformeR \cite{liu2021cptr} (CPTR) model, which is a full Transformer network without any convolutional operation in the encoder. Unlike previous models, the CPTR model uses the raw image, divides it into N patches, reshapes the patches into vectors and learns features from them, with positional encoding, using the transformer encoder. 
\\
\\
\textbf{Chest-Xray Report Generation}
\\
Recent approaches to automatic radiology/CXR image report generation are predominantly based on deep learning and typically adopt an image captioning approach, utilising a combination of convolution and recurrent neural networks. For example, Jing et al\cite{jing2017automatic}, used a pre-trained VGG-19 model to learn visual features, which are then used to predict relevant tags for any given chest X-ray. These tags become semantic features in the network, which alongside visual features, are fed into a Co-Attention Network. Subsequently, a Hierarchical LSTM uses the context vector provided by the Co-Attention Network to generate a description for the given X-ray. While the model achieved promising results, there were concerns around the repetitive sentences in reports and the inconsistency in generating text for the same patient during inference, where the network produced different outcomes. Another study\cite{xue2018multimodal} addressed this issue by enhancing the pre-trained Resnet-152 encoder using multi-view content and incorporating a sentence decoder to generate a report. Multi-view approach mitigated the problem of report variability for the same patient. Additionally, they used the first predicted sentence as a joint input alongside image encoding, ensuring consistency in the results.

An alternative proposal\cite{yuan2019automatic} was posited, wherein they chose to pre-train their multi-view encoder from randomly initialised weights using the CheXpert dataset \cite{irvin2019chexpert}, rather than relying on pre-trained models from ImageNet. To improve the efficacy of the decoder, they extracted normalised medical concepts from radiology reports using Semrep \footnote{https://semrep.nlm.nih.gov/}. These extracted concepts were embedded into the decoder in two ways: 1) they concatenated the concept embedding with the encoder output before feeding to the decoder, providing explicit semantic information, and 2) they used a concept-aware attention mechanism that attends over the embedded concepts when generating words, enabling the decoder to focus on relevant medical terms. By enriching the decoder with explicit medical concept knowledge, their model could generate reports with more accurate and meaningful terminology aligned with the clinical finding descriptions. Medical concepts were also employed in another study\cite{yang2020automatic}, where, the authors introduced  a reinforcement learning-based reward for concept extraction to obtain more accurate and precise concepts. This reward encouraged the model to extract concepts that have a higher likelihood of being mentioned in the radiologist's report. Compared to previous work that extracted concepts without optimisation, their approach achieved higher precision and coverage of concepts that radiologists tend to use in real reports. However, the authors acknowledged their generated reports still lacked some descriptive informativeness compared to ground truth. Their analysis found that only extracting concepts from previous reports limits the diversity of expressions in generated reports. 

A study\cite{singh2021show} highlighted the frequent discrepancy between the formats of normal and abnormal radiology reports, with abnormal reports indicating the suspicion of an abnormality or pathology. To address this issue, they categorised reports as either normal or abnormal based on the content of the report text. For the report generation process, they adopted a two-stage approach. First, they generated the 'Findings' section, which describes the visual observations made by radiologists during the examination of medical images. This was achieved by leveraging visual features extracted from a CNN model and relevant report texts. Subsequently, they summarised the generated 'Findings' to produce the 'Impression' section, which provides a summarised interpretation of the radiologist's observations. A key contribution of their study was showing that conditioning the text generation on the report type (normal or abnormal) improved the clinical validity and alignment with real radiology reporting practices.

Recent studies have also  capitalised on Transformer models for medical report generation, after having achieved success in text generation based on non-linguistic representation. One such study \cite{xiong2019reinforced} constructed a hierarchical Transformer model which features a novel encoder capable of extracting regions of interest from the original image via a region detector, and subsequently, utilising these regions to obtain visual representations. Additionally, another study\cite{chen2020generating} introduced a medical report generator utilising a memory-driven Transformer. They proposed a relational memory (RM) module to retain knowledge from previous cases, thereby enabling the generator model to remember similar reports when generating current reports. Another study \cite{nooralahzadeh2021progressive} proposed a progressive Transformer-based framework for report generation, which generates high-level context from the given X-ray and then employs the Transformer architecture to convert this context into a radiology report. This model comprises a pre-trained CNN as a visual backbone, a mesh-memory Transformer\cite{cornia2020meshed} as a visual-language model, and BART \cite{lewis2019bart} as a language model.

With increasing interest in this application domain, studies have become more attentive to the distinctions between image captioning and report generation tasks. As a result, researchers have begun to develop more knowledge-informed networks tailored specifically to the task of image-guided radiology report generation. The paper \cite{wang2022inclusive} introduces a task-aware framework that is designed to be adaptable to different imaging types and medical scenarios. It prioritises understanding specific diagnostic tasks related to various medical conditions, ensuring accurate and contextually relevant report generation. Another study \cite{yang2022knowledge} highlights the significance of both input-independent general medical knowledge and input-dependent specific contextual information in generating accurate chest radiology reports. They proposed a knowledge-enhanced method that leverages these information along with visual features to improve the quality and accuracy of generated reports for chest X-rays. Recently, another study \cite{wu2023multimodal} introduced a technique called multi-modal contrastive learning, which aims to enhance the synergy between different modalities of data. By leveraging contrastive learning, the proposed method aligns and embeds visual and textual representations in a shared space, facilitating the generation of more informative and accurate radiology reports.
\\
\\
\textbf{Fusion Strategies}
\\
Data fusion refers to the integration of different data modalities that provide separate perspectives on a problem to be addressed, and using multiple modalities has the potential to decrease the number of errors compared to approaches that only use one type of data\cite{stahlschmidt2022multimodal}. Deep learning fusion strategies can be broadly classified into three categories: early fusion, late/decision fusion, and hybrid/joint fusion. In the process of early fusion, the original or transformed features are combined at the input level before being fed into a single model that can handle all the information. There are various methods of joining data, but early fusion commonly involves concatenation or pooling. In late fusion, the input data is processed independently through separate networks. The outputs from these networks are then combined at a later stage to form a joint decision. Late fusion strategies learn modality-specific features separately and then integrate them downstream in the model (e.g. just before the prediction/output layer). Lastly, joint fusion involves combining the features extracted from different modalities at different stages of the network architecture.

Within the medical imaging field, the utilisation of multi-modal data fusion approaches has the potential to enhance performance in addressing complex tasks that exceed the capabilities of a single imaging modality. Concentrating on chest X-ray modality, multiple tasks such as image classification, image retrieval, and modality translation have leveraged data fusion strategies. For example, one study\cite{wang2018tienet} introduced a CNN-RNN architecture called the text-image-embedding network (TieNet) to extract discriminative representations of both chest radiographs and their accompanying reports by combining visual and textual information through joint fusion. The experimental results indicate that TieNet's multimodal approach outperforms its unimodal counterpart in multi-label disease classification. Another study \cite{chauhan2020joint} employed a semi-supervised approach to train the network on chest radiographs and associated radiology reports to evaluate the severity of pulmonary edema. This study demonstrated that joint learning of image-text representations enhances the performance of models designed to predict the severity of pulmonary edema, compared with supervised models that relied solely image-derived features. A paper \cite{hayat2022medfuse} discussed the challenge of integrating data from different sources in healthcare due to asynchronous collection of modalities. They proposed an LSTM-based fusion module, called MedFuse, that accommodates uni-modal and multi-modal input for mortality prediction and phenotype classification tasks. In contrast with intricate multi-modal fusion techniques, MedFuse yields considerably better performance on the fully paired test set, furthermore, it demonstrates robustness when dealing with the partially paired test set, which includes instances of missing chest X-ray images.
\\
\\
This paper introduces a novel multi-modal deep neural network for generating radiology reports based on quantitative image analysis, patient demographic information, and other clinical data collected during a patient's stay. The main objective is to generate a consistent and comprehensive report that adheres to the format used in real-world clinical practice. Our contributions are as follows:
\\
\\
To the best of our knowledge, this is the first attempt to generate an automatic CXR report by integrating patient information and clinical data obtained from CXR exams, which is not typically included in radiology reports. We propose a novel conditioned cross-multi-head attention module to fuse structured data, unstructured text and visual information. Additionally, we employed human evaluation and clinical semantic similarity measurement(Bio-ClinicalBERT Score\cite{alsentzer2019publicly}) alongside word-overlap metrics to improve the depth of quantitative analysis. Our experiments demonstrate that the incorporation of additional data not explicitly stated in the report enhances the model's performance. Our proposed model achieves the best performance on the ROUGE-L metric \cite{lin2004rouge} when compared to similar state-of-the-art studies.
\section{Materials and methods}
\begin{figure*} [t]
  \centering
  \includegraphics[width=0.85\textwidth,height=8cm]{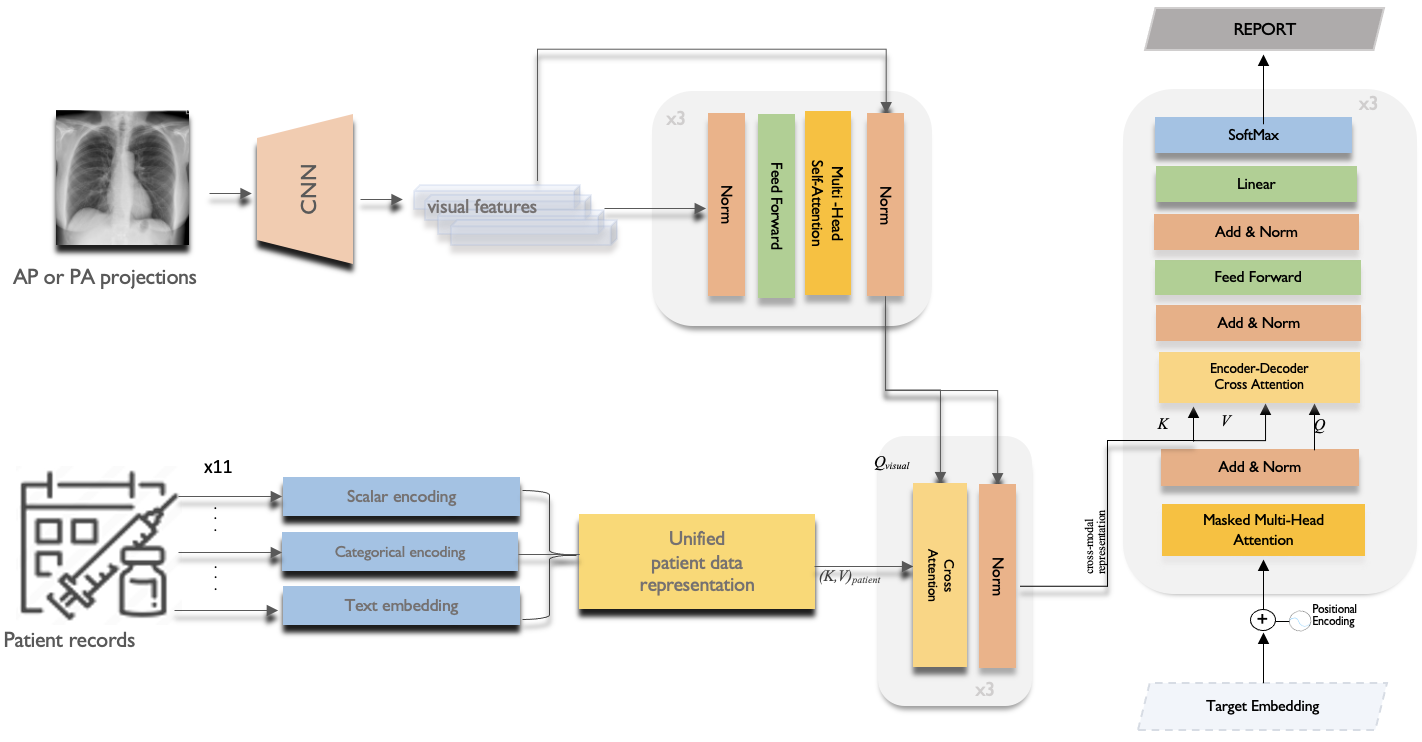}
  \caption{The overall multi-modal data fusion with cross-attention framework of the proposed CXR report generation model.}
  \label{fig:multi-modal-framework}
\end{figure*}
\subsection{Data}
The dataset used in this study was created by leveraging three openly accessible databases, namely MIMIC-CXR, MIMIC-IV, and MIMIC-IV-ED. MIMIC-CXR (version 2.0) encompasses a vast collection of 377,110 CXR images captured from multiple views, together with 227,835 de-identified radiology reports, pertaining to 63,473 patients. Each report contains several sections, such as 'examination', 'indication', 'technique', 'comparison', 'findings', and 'impressions'. Meanwhile, MIMIC-IV (version 2.0) comprises de-identified patient data, including characteristics like age, gender, ethnicity, and marital status, extracted from individuals who were admitted to Beth Israel Deaconess Medical Center (BIDMC). Furthermore, MIMIC-IV-ED (version 2.2) is an extensive database of emergency department (ED) admissions at the BIDMC between 2011 and 2019, which contains detailed clinical information, including diagnosis, medication, triage, and vital signs.

Each of the databases comprises distinct tables containing varying details related to a patient's hospitalisation. An individual patient is assigned a unique identifier, referred to as the subject id. However, since a single patient might have multiple hospitalisations, or a single stay may generate several records, linking these databases using subject id proved unfeasible. Moreover, as the aim is to generate an accurate report, it is imperative that non-imaging data be collected within the same time frame as the chest x-ray. Consequently, we resorted to record linkage between MIMIC-CXR and MIMIC-IV-ED databases and extracted data only if the patient was in the ED while the report was being generated and did not leave during that period. After performing data cleaning procedures; excluding missing data, filtering the images to only include anteroposterior and posteroanterior projections, keeping only one study if there is more than one records, and removing replications, the resulting dataset contains 65813 entries and 11 features including acuity level, oxygen saturation, heart rate, respiratory rate, systolic blood pressure, diastolic blood pressure, temperature, patient's chief complaint, ICD title, gender and ethnicity. 
\\
\\
One challenge with the dataset for this task is its biases, particularly, its skewed distribution towards normal cases and the presence of numerous identical reports for different patients. To minimise these issues,  we selected a subset of 65813 entries, by identifying and cataloging unique medical reports, ensuring that each distinct group was represented in the curated dataset to a similar extent. This subset consisted of 3000 total samples, which we further divided into a training set comprising 2100 datapoints and a validation set comprising 900 datapoints. Subsequently, we evaluated the performance of our models on a holdout test set comprising 1173 unseen examples. As there is currently no comparable comprehensive dataset encompassing similar non-imaging data, we exclusively employed this specific dataset to train and assess the proposed approach.

\subsection{Feature Extraction and Pre-processing}
This section describes the pre-processing and encoding of different data modalities used in this study. The main objective is to bridge the gap between the data used in the study and the data typically encountered in medical practice while minimising potential biases that may arise.

\textbf{\textit{Image Data}}
\\
Each image went through resizing to 299 pixels x 299 pixels, followed by min-max normalisation to scale the intensity values to a range of 0 to 1. The process of obtaining the representation of each image can be described as a two-step procedure. In the first step, the EfficientNet model is utilised as the base model to extract the visual features of the image. In the second step, this feature vector is employed as the input for a transformer-based encoder which extracts higher-level features and fuses this information with clinical and non-clinical(demographic) data. A detailed explanation of the fusion process can be found in section 3.3.

\textbf{\textit{Clinical: Non-imaging Data}}
\\
This study exclusively employed clinical data that clinicians considered during patient evaluations, wherein, a chest X-ray examination was conducted if any disease/abnormalities were suspected. These data included heart rate, respiratory rate, oxygen saturation, temperature, level of acuity (severity), primary symptoms or complaints, as well as known or suspected diseases. 

The acuity level of a patient is determined based on the triage assessment, and an integer value, \textit{between 1 and 5}, is assigned to each case where 1 indicates the least severe and 5 is the most severe. The higher acuity levels are typically associated with the presence of abnormalities in the patient's case, therefore, utilising the acuity level may assist the network in determining normal and abnormal cases while generating the report.

The integer-based variables including oxygen saturation, heart rate, respiratory rate, systolic blood pressure (sbp), and diastolic blood pressure (dbp) were initially treated to remove outliers. Subsequently, the values have been normalised within the range of 0 to 1, based on their respective minimum and maximum values. As for temperature data, a conversion to Fahrenheit scale was performed, and similar to the integer-based variables, the values were normalised between 0 and 1. 

The text-based variables, namely the chief complaint and ICD title variables, are initially processed by converting characters to lowercase and removing unnecessary punctuation, such as commas, periods, and newline characters, utilising regular expressions. Consecutive periods are condensed into single spaces, and double periods are substituted with single spaces, contributing to a more consistent text format. The resultant text undergoes further standardisation by substituting shorthand phrases or abbreviations with their corresponding full-text counterparts. For example, 'cp' is replaced with 'chest pain', 'sob' or 'shortness of breath' is replaced with 'dyspnea' and so on. Standardisation also includes converting phrases like "chest pain, dyspnea" into "chest pain and dyspnea" as well as fixing typos and pluralisation issues such as changing "fevers" to "fever." 

\textbf{\textit{Non-clinical Data}}
\\
In addition to clinical data, patient records often include non-clinical metadata that can provide valuable insights. This study concentrates on two commonly collected non-clinical variables: gender, and ethnicity. These variables have been demonstrated to have an impact on health outcomes \cite{aksoy2023radiology} and are therefore of particular interest in this paper.

As the gender data is already in binary format, the only necessary pre-processing step was to convert the data to a numerical representation by replacing 'Male' with 0 and 'Female' with 1. 

The ethnicity data was initially categorised into 5 broad groups consisting of the most frequently occurring values and this initial categorisation slightly improved model performance. The data was then categorised in a more granular fashion into 9 groups: White, African American, Hispanic/Latino,  Black, Asian, White/European, Russian, Other, and Unknown. We hypothesised that employing these more detailed ethnicity categories would enable more accurate report generation. Subsequently, the categorical ethnicity data was mapped to integer values and reshaped into a 2D array to allow for input into the encoder.

\subsection{Multi-modal Data Fusion Details}
The scalar patient data, comprising attributes like heart rate, oxygen saturation, respiratory rate, blood pressure, temperature, acuity level and gender, is concatenated to form a continuous representation. This continuous data is then passed through a dense layer to produce a scalar output.

Each ethnicity group variable is transformed using the one-hot encoding (Equation \ref{eq:ethnicty}), resulting in a matrix where each individual's ethnicity is represented as a binary vector.
\begin{equation} \label{eq:ethnicty}
X_{\text{eth}} = [\delta(\text{eth}, 1), \delta(\text{eth}, 2), \ldots, \delta(\text{eth}, 9)].
\end{equation} 
\begin{center}
where the function $\delta(i, j)$ is defined as
\end{center}
\[
\delta(i, j) = \begin{cases}
    1 & \text{if } i = j, \\
    0 & \text{if } i \neq j.
\end{cases}
\]
\\
The chief complaint and ICD title data consist of text sequences with varying lengths and vocabulary sizes. Therefore, these data are separately embedded using the following embedding technique before being further processed through dense layers.
\\
\\
Let,
\begin{minipage}[t]{0.45\textwidth}
\[
x_{text\_data} \in \mathbb{Z}^{N \times M} \text{~---~} \text{2D input tensor of indices}
\]
\[
V_{text\_data} \text{~---~} \text{vocabulary size}
\]
\[
E \text{~---~} \text{embedding dimension}
\]
\[
W_{\text{emb}} \in \mathbb{R}^{V_{text\_data} \times E} \text{~---~} \text{embedding weight matrix}
\]
\end{minipage}
\hfill
\begin{minipage}[t]{0.45\textwidth}
Then,
\begin{equation*} 
f_{\text{embed}}(x_{\text{text\_data}}) \in \mathbb{R}^{N \times M \times E}
\end{equation*}
\begin{equation}
f_{\text{embed}}(x_{\text{text\_data}})_{i,j,k} = W_{\text{emb}}[x_{\text{text\_data}_{i,j}}, k]
\label{eq:fembed}
\end{equation}
where
\end{minipage}

\begin{equation*}
\begin{split}
f_{\text{embed}}(x_{\text{text\_data}})_{i,j,k} \text{---} &\text{embedding vector} \\
&\text{for token at position}~(i, j)
\end{split}
\end{equation*}

\begin{equation*}
x_{\text{text\_data}_{i,j}} \text{---} \text{integer index of token at position}~(i, j)
\end{equation*}

\begin{equation*}
\begin{split}
W_{\text{emb}}[x_{\text{text\_data}{i,j}}, k] \text{---} k\text{-th value from row of}\\
~W{\text{emb}}\text{~for index}~x_{\text{text\_data}_{i,j}}
\end{split}
\end{equation*}

\begin{align*}
X_{\text{chief\_embed}} &= f_{\text{embed}}(x_{\text{chief\_data}}) \\
X_{\text{icd\_embed}} &= f_{\text{embed}}(x_{\text{icd\_data}})
\end{align*}
Where:
$x_{\text{chief\_data}}$ and $x_{\text{icd\_data}}$ are the respective input indices tensors and $f_{\text{embed}}$ is the embedding function defined in Equation \ref{eq:fembed}.

After feature extraction and transformation of patient data inputs, the representations are concatenated into a unified patient representation vector. The processed scalar data output \(X_{\text{scalar}} \in \mathbb{R}^{N \times M_{\text{scalar}}}\), one-hot encoded ethnicity output \(X_{\text{eth}} \in \mathbb{R}^{N \times M_{\text{eth}}}\), and embedded chief complaint and ICD title outputs \(X_{\text{chief\_embed}}, X_{\text{icd\_embed}} \in \mathbb{R}^{N \times M \times E}\) are concatenated for each patient giving:
\begin{equation}
\begin{aligned}
X_{\text{patient}} = \text{Concatenate}&(X_{\text{scalar}}, X_{\text{eth}}, X_{\text{chief\_embed}}, \\
& X_{\text{icd\_embed}};   \text{axis} = 1)
\end{aligned}
\end{equation}

Where the \text{Concatenate()} operation joins the input tensors along the specified dimension, in this case, \text{axis}=1, yielding:
\begin{equation}
\begin{aligned}
X_{\text{patient}} &= \begin{bmatrix} X_{\text{scalar}} & X_{\text{eth}} & X_{\text{chief\_embed}} & X_{\text{icd\_embed}} \end{bmatrix} \\
&\in \mathbb{R}^{N \times (M_{\text{scalar}} + M_{\text{eth}} + 2M \times E)}
\end{aligned}
\end{equation}

The resulting \(X_{\text{patient}}\) contains a unified representation of each patient's data for further use, combining structured scalar variables, categorical encodings, and semantically rich embedded features into a single vector. This concatenation enables the joint modeling of heterogeneous data types into an integrated patient representation. 

Then, an EfficientNetB0 CNN backbone \cite{tan2019efficientnet} pre-trained on ImageNet extracts features from 299x299x3 RGB input images. The CNN outputs N x D image embeddings, where N is batch size and D is the feature dimension. This 1280-length visual feature vector is transformed via layer normalisation and a dense layer to refine the image representation. Before starting to fusion operation, multi-headed self-attention (Equation \ref{eq:mha}) is then applied to enable the model to jointly focus on different positions in the image via parallel heads. The self-attention outputs are then added to the original embedded image via residual connection, and normalised by a layernorm layer. 
\begin{equation} \label{eq:attention}
    Attention(Q, K, V) = softmax(\frac{QK^T}{\sqrt{d_k}})V
\end{equation}
where $\sqrt{d_k}$ is the dimension of the key vector $k$ and query vector $q$ 
\begin{equation} \label{eq:mha}
MultiHead(Q, K, V) = Concat(head_1, ..., head_h)W^O
\end{equation}
where
\begin{equation*}
head_i = Attention(Q W^Q_i, K W^K_i, V W^V_i)
\end{equation*}
The final output image embedding is further contextualised with information from the entire set of patient data via a cross-attention mechanism. Specifically, final image embedding is used as the query matrix \(Q\), while the unified patient data embedding \(X_{\text{patient}}\)  serves as the key and value matrices \(K\) and \(V\). 

\[Q \in \mathbb{R}^{N \times D}\]
\[X_{\text{patient}} \in \mathbb{R}^{N \times D}\]
\[K = V = X_{\text{patient}}\]
where \(N\) is the batch size, and \(D\) is the embedding dimension.

Multi-headed scaled dot-product attention is again applied between \(Q\) and \(K\) to obtain attention weights representing the relevance of each part of the patient data to each part of the image. The weighted value matrices are concatenated and projected to obtain the cross-attention outputs allowing the model to condition each part of the image embedding on relevant unified patient representation. The cross-attention outputs are residually connected and normalised in a similar manner via element-wise addition with the output image embedding from the previous self-attention block and layernorm.

We adopt the canonical Transformer decoder architecture and perform attention over the encoded cross-modal representation obtained from the encoder. This enables each decoded output embedding to be conditioned on the relevant semantic concepts and modalities from the encoder, facilitating more effective fusion and reasoning across modalities.

\section{Experiments and Analysis}

\begin{table*}[t] 
\captionsetup{skip=10pt}
\begin{center} 
\resizebox{\textwidth}{!}{%
\begin{tabular}{lcccccccc} 
\toprule
Method & B\_1 & B\_2 & B\_3 & B\_4 & R\_L & BS$_{F1}$ & Bio-CBS$_{F1}$ & \\ 
\midrule
Baseline & 0.326 & 0.205 & 0.138 & 0.084 & 0.301 & 0.192 & 0.787 \\
Singular02Sat & 0.343 & 0.222  & 0.151  & 0.096  & 0.321  & 0.199  & 0.789\\
TextFusion & 0.326 & 0.209 & 0.141 & 0.862 & 0.307 & 0.181 & 0.784\\
ScalarFusion & 0.343 & 0.219 & 0.145 & 0.090 & 0.320 & 0.198 & 0.786 \\
AllDataFusion & \textbf{0.351} & \textbf{0.231} & \textbf{0.162} & \textbf{0.107} & \textbf{0.331} & \textbf{0.218} & \textbf{0.794} \\ 
\bottomrule
\end{tabular}%
}
\caption{Quantitative Comparison of Fusion Methods: Performance Evaluation Across Multiple Metrics.
\\B\_n for BLEU-n, R\_L for ROUGE-L, BS$_{F1}$ for BERT Score F1Score and CBS$_{F1}$ for Bio-ClinicalBERT Score F1Score.}
\label{tab:results} 
\end{center}
\vspace{10pt} 
\end{table*}

\subsection{Experimental Setup}
The model undergoes training through a custom loop that involves the following key steps: data retrieval, image embedding, encoding of clinical and non-clinical data, calculation of loss and accuracy, computation of gradients, weight updating, and tracker adjustment. 

The standardised sequence lengths are 43 tokens for reports, 2 tokens for chief complaints, and 6 tokens for ICD codes. The vocabularies contain over 6,000 unique tokens for radiology reports, and over 3,000 tokens each for chief complaints and ICD codes. The vocabulary size and fixed sequence length were determined based on the complete dataset, not just the balanced subset of 3,000 samples used for training and validation. Both image features and text tokens are represented using 512-dimensional embeddings. The Transformer encoder and decoder layers include feed-forward networks with 512-dimensional units each, and Transformer layers utilize multi-headed attention with 3 attention heads. During training, a batch size of 64 is employed, and training proceeds for 100 epochs with early stopping triggered by validation loss stagnation over 5 epochs. 

The model's training employed the Adam optimiser with a learning rate of 3e-4 and linear warmup for the first 500 steps. After the warm-up phase, the learning rate remains constant, stabilising training and facilitating effective model fine-tuning. Loss is calculated using the Sparse Categorical Cross-Entropy loss function defined in Equation \ref{eq:lossfunction}, and accuracy is assessed by matching predicted tokens with true tokens.
Let:
$y_{\text{true}_i}$ be the ground truth for the radiology reports.
$y_{\text{pred}_i}$ be the output from our report generation model.
The equation for cross-entropy loss for each report without reduction is given by:
\begin{equation} \label{eq:lossfunction}
\text{loss}_i = - \sum (y_{\text{true}_i} \cdot \log((y_{\text{pred}_i})))
\end{equation}

Where:
$i$ represents the index of the report.
$y_{\text{true}_i}$ is the ground truth for report $i$.
$y_{\text{pred}_i}$ is the generated report for index $i$.
This loss calculation is performed for each report separately, without any reduction.

\subsection{Evaluation Metrics}
To evaluate the linguistic quality of the generated radiology reports, we computed several automatic evaluation metrics comparing the generated text to the reference reports. First, BLEU-1 to BLEU-4 scores \cite{papineni2002bleu} were calculated to assess n-gram precision for unigrams up to 4-grams. Higher BLEU scores indicate greater local word-level similarity between the generated and reference texts. Second, the ROUGE-L score was used to measure the longest common subsequence, assessing the quality of the generated text in terms of recall and precision.

Additionally, we evaluated semantic similarity using the BERT Score \cite{zhang2019bertscore}. However, generic BERT representations may not fully capture domain-specific conceptual information needed for clinical text generation. To address this, we initialise BERTScore with the Bio-ClinicalBERT embeddings \cite{alsentzer2019publicly} trained on scientific text and clinical notes. By plugging these contextual embeddings into the BERTScore framework, we obtain a domain-adapted evaluation metric that emphasises clinical conceptual similarity. Specifically, the F1 component now computes semantic textual similarity using Bio-ClinicalBERT's clinical embeddings rather than generic BERT. These metrics providing a more nuanced assessment of meaning compared to strict n-gram matching. The BERT-based metrics can capture whether the generated reports convey clinically coherent descriptions despite differing word usage compared to the reference. Collectively, these automated evaluation metrics quantify linguistic similarity at both word-level, sentence-level, and semantic meaning levels.
\subsection{Quantitative Results}

In this study, we leverage 11 additional clinical features along with chest x-rays to generate more accurate and informed radiology reports. The baseline model only employed chest X-ray images as input to generate corresponding reports, serving as our benchmark reference where the sole source of information was the visual data.

To analyse the impact of individual features (Table \ref{tab:singularmodels}), we evaluate a Singular02Sat model that incorporates just a single additional feature – oxygen saturation (O2Sat) – with the chest x-ray. O2Sat is chosen as it demonstrated the highest performance among the singular models. The TextFusion model explores using only the text-based features of primary symptoms and ICD diagnostic titles as inputs alongside the chest x-ray. Furthermore, a ScalarFusion model fuses multiple scalar features, each of which, when integrated individually through their respective singular models, demonstrates good enhancements over the baseline model.. These scalar features include oxygen saturation (o2sat), diastolic blood pressure (dbp), temperature (temp), patient acuity (acuity), and gender.

At the core of our study, the AllDataFusion model takes a holistic approach by fusing all available and relevant data points in the dataset. This multi-modal fusion aims to effectively incorporate the diverse sources of information at hand, including chest x-ray images, structured clinical data, and unstructured text notes.

Table~\ref{tab:results} presents a quantitative comparison based on the performance across multiple evaluation metrics. The metrics utilized for the assessment include BLEU-n (B$_1$ to B$_4$), ROUGE-L (R$_L$), BERT F1Score (BS$_{F1}$), and Bio-ClinicalBERT F1Score (Bio-CBS$_{F1}$). The Singular02Sat method displayed notable improvements across multiple metrics compared to the baseline, while the TextFusion and ScalarFusion methods showcased marginal increases. The AllDataFusion method emerged as the top performer, showing substantial enhancements in various metrics, highlighting the benefits of multi-modal fusion.
\\
\\
Among the test samples, approximately 33\% of them have  BLEU-1 scores between 0.1 and 0.3, around 54\% have scores between 0.3 and 0.5, about 11\% have scores between 0.5 and 0.7, and a mere 0.26\% have scores between 0.7 and 1, indicating high similarity. Our BLEU-1 results exhibit strong concordance with existing report generation literature, which has established scoring norms averaging 0.3 to 0.4 for this metric.We conducted a comparative analysis of our model against relevant state-of-the-art models (\cite{yang2022knowledge, wu2023multimodal, nooralahzadeh2021progressive, wang2022inclusive}), referencing the results documented in their published literature. When considering ROUGE-L score, which reflect model's ability to capture document-level linguistic coherence, our approach excelled in this aspect, achieving a ROUGE-L score of 0.331, standing out as the highest score across all models. This suggests that our model excels in capturing the long-range linguistic context of medical reports.

\begin{table}[H]
    \centering
    \small
    \begin{tabularx}{0.5\textwidth}{l*{5}{X}}
        \toprule
        Singular Model & B\_1 & B\_2  & BS$_{F1}$  & Bio-CBS$_{F1}$ & R\_L \\
        \midrule
        ETHNICITY & 0.328 & 0.212 & 0.174 & 0.782 & 0.321 \\
        HEARTRATE & 0.333 & 0.213 & 0.170 & 0.786 & 0.295 \\
        ICD & 0.328 & 0.207 & 0.186 & 0785 & 0.301 \\
        RESPRATE & 0.329 & 0.213 & 0.185 & 0.788 & 0.309 \\
        SBP & 0.336 & 0.219 & 0.197 & 0.786 & 0.319 \\
        DBP & 0.338 & 0.220 & 0.188 & 0.784 & 0.322 \\
        O2SAT & \textbf{0.343} & 0.222 & 0.199 & \textbf{0.789} & 0.321 \\
        ACUITY & 0.342 & 0.222 & \textbf{0.200} & \textbf{0.789} & 0.309 \\
        TEMPERATURE & 0.336 & 0.219 & 0.199 & \textbf{0.789} & 0.326 \\
        GENDER & 0.341 & \textbf{0.224} & 0.197 & 0.787 & \textbf{0.331} \\
        CHIEFCOMPLAINT & 0.326 & 0.212 & 0.185 & 0.785 & 0.302 \\
        \bottomrule
    \end{tabularx}
    \caption{Singular model results}
    \label{tab:singularmodels}
\end{table}

\subsection{Qualitative Results}
\begin{figure*} [t]
  \centering
  \includegraphics[width=0.9\textwidth,height=8cm]{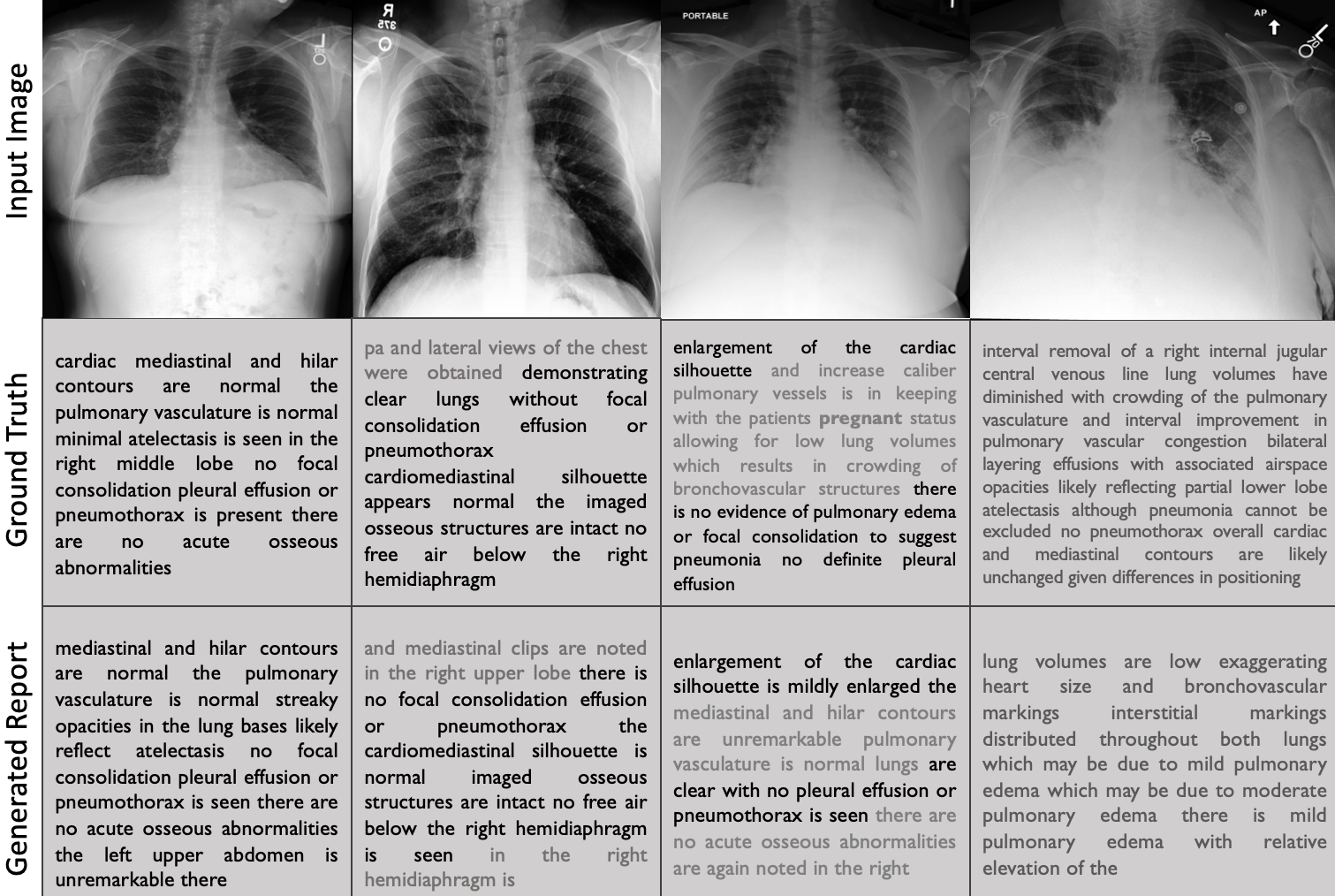}
  \caption{GT and GR report from the proposed AllDataFusion CXR report generation model.}
  \label{fig:QualitativeResults}
\end{figure*}

Figure \ref{fig:QualitativeResults} shows the ground truth and generated report from our Alldatafusion model. For a better assessment of the results, we illustrated the samples that show accurate prediction, different expressions, missing and false arguments, and false prediction.

The results show promise in producing reports that capture many of the key findings described in the ground truth reports. In all cases, the order of findings aligns with the reports written by the radiologists and the generated reports are structurally correct. The results reveal a generally positive alignment in terms of language and grammar, however, some of the generated reports exhibit repeated words or phrases, which can affect the overall coherence. Additionally, the usage of "and" at the beginning of sentences and concluding paragraphs with "the" or "is" reveals grammatical inconsistencies. 

The model accurately identifies normal cardiac, mediastinal, and hilar contours when present in the ground truth. It also reliably notes the presence or absence of abnormalities like pulmonary edema, pleural effusion, focal consolidation, pneumonia, and pneumothorax which are crucial in radiology analysis. In some cases, the generated reports exhibit a reduced level of detail compared to the ground truth, omitting certain specific observations. In the first sample, the model missed the right middle lobe atelectasis that was noted in the ground truth. In the second sample, the model hallucinated mediastinal clips not present in the ground truth or image. The third sample shows that the model did not fully capture the enlarged cardiac silhouette and vessels described in the ground truth. In the last sample, the model missed details about the interval removal of a central venous line and differences in positioning compared to a prior exam that provided important clinical context.

Overall, these qualitative results demonstrate good progress for the radiology report generation model, with accurate high-level identification of key findings, but also room for improvement in capturing more nuanced details and clinical context. While the baseline model was capable of providing results, it did not exhibit the same level of detail and accuracy as the enhanced model. Continued refinement of the model will be important to ensure accurate detection and description of abnormalities.

\subsection{Radiologist Evaluation Results}
We evaluated the model using 158 randomly selected samples from the unseen test set, covering diverse medical conditions reflecting the full distribution. A board-certified radiologist assessed three criteria: language fluency, content selection, and correctness of abnormal findings (AF). For language, the radiologist evaluated sentence structure, terminology, and overall clarity. For content, they compared the report's level of detail, key findings, and image coverage to the true findings. They assessed accuracy of abnormal findings by comparing to the true conditions. The radiologist assigned 1-5 scores and noted preference between reports.
This methodology enabled quantitative and qualitative assessment of language generation, content selection, and diagnostics. The radiologist also noted that while performing well overall, some shortcomings were observed. The model often missed surgical materials like catheters and clips and it fails to capture anomaly variations when the patient is inclined to the right or left. Sensitivity to bone lesions was lacking, overlooking non-urgent findings like scoliosis.  However, it's worth noting that these are not extensively covered in the ground truth as well. For normal x-rays, it occasionally included non-definitive elements. While these additions may be accurate, there is a slight possibility that they may not be. This evaluation methodology provided valuable insights into model strengths and areas needing improvement.

\begin{table}[H]
    \centering
    \small
    \begin{tabularx}{0.4\textwidth}{l*{2}{X}}
        \toprule
        Language Fluency & Content Selection & Correctness of AF\\
        \midrule
         4.24 & 4.12 & 3.89 \\
        \bottomrule
    \end{tabularx}
    \caption{Radiologist Evaluation Results on a 1-5 Scale}
    \label{tab:radiologist_results}
\end{table}

\section{Discussion and Conclusion}

In this study, we adopted a comprehensive approach to enhance the precision and clinical relevance of radiology reports generated in conjunction with chest X-ray images. We achieved this by incorporating cutting-edge network components and drawing inspiration from state-of-the-art methods, such as the transformer architecture and multi-modal data fusion techniques. In the section pertaining to data selection, we ensured the inclusion of data only within the time frame of radiology report generation. This was done to closely replicate the clinical pathway. Additionally, we aimed for a balanced representation of each type of report in our sample selection to counteract potential biases skewed towards normal cases. Although this approach resulted in a smaller dataset compared to existing literature, it was a crucial step for preventing biased results and validating the results in real clinical settings.

Recent literature highlights the potential of multi-modal learning techniques in advancing the quality of automated radiology report generation. However, a majority of medical report generation models primarily focus on target reports within specific information or incorporate image findings as supplementary inputs. Given that Chest X-rays present three-dimensional objects in a two-dimensional form, some valuable information is lost, leading to semantic gaps in the data provided to the network. Furthermore, radiologists possess more data beyond images during report creation. To address this limitation, we bridged the semantic gap between vision and language models by capturing uncodified information essential to the diagnosis process. We achieved this by introducing an ensemble of 11 supplementary features in conjunction with the chest X-ray data. These features were thoughtfully selected to enhance both accuracy and clinical insight in the generated reports.

The results indicate that incorporating non-imaging clinical and non-clinical data positively impacts the quality of the generated reports. Our ablation study further demonstrates that providing all data simultaneously yields higher accuracy compared to using individual data components separately. This finding suggests that introducing data with no significant standalone impact on the network alongside others leads to improved results. In light of these outcomes, we recommend that the broader research community include additional metadata for enhanced model performance.


\acknowledgments 
This study is fully sponsored by the Turkish Ministry of National Education. 
\bibliography{report} 
\bibliographystyle{reportbib} 

\end{multicols}

\end{document}